\documentclass{article}

\PassOptionsToPackage{numbers, compress}{natbib}

\usepackage[preprint]{neurips_2023}

\usepackage[utf8]{inputenc} 
\usepackage[T1]{fontenc}    
\usepackage{hyperref}       
\usepackage{url}            
\usepackage{booktabs}       
\usepackage{amsfonts}       
\usepackage{nicefrac}       
\usepackage{microtype}      
\usepackage{xcolor}         

\usepackage{graphicx, caption} 
\usepackage{svg}
\usepackage{caption}
\usepackage{subcaption}
\usepackage{amsmath, amssymb, amsfonts, amsthm}

\usepackage[export]{adjustbox}[2011/08/13]

\title{Tryage: Real-time, Intelligent Routing of User Prompts to Large Language Models}

\author{Surya Narayanan Hari \\
Department of Biology and Biological Engineering \\
California Institute of Technology\\
 \texttt{shari@caltech.edu}
\AND
Matt Thomson\\
Department of Biology and Biological Engineering\\
Program in Computational and Neural Systems\\
California Institute of Technology\\
 \texttt{mthomson@caltech.edu}}

\begin{document}
\maketitle
\begin{abstract}
The introduction of the transformer architecture and the self-attention mechanism has led to an explosive production of language models trained on specific downstream tasks and data domains. With  over $200,000$ models in the Hugging Face ecosystem, users grapple with selecting and optimizing models to suit multifaceted workflows and data domains while addressing computational, security, and recency concerns.  There is an urgent need for machine learning frameworks that can eliminate the burden of model selection and customization and unleash the incredible power of the vast emerging model library for end users. Here,  we propose a context-aware routing system, Tryage, that leverages a language model router for optimal selection of expert models from a model library based on analysis of individual input prompts. Inspired by the thalamic router in the brain,  Tryage employs a perceptive router to predict down-stream model performance on prompts and, then,  makes a routing decision using an objective function that integrates performance predictions with user goals and constraints  that are incorporated through flags (e.g., model size, model recency). Tryage allows users to explore a Pareto front and automatically trade-off between task accuracy and secondary goals including minimization of model size, recency, security, verbosity, and readability. Across heterogeneous data sets that include code, text, clinical data, and patents, the Tryage framework surpasses Gorilla and GPT3.5 turbo in dynamic model selection identifying the optimal model with an accuracy of $50.9 \%$ , compared to $23.6\%$ by GPT 3.5 Turbo and $10.8\%$ by Gorilla. Conceptually, Tryage demonstrates how routing models can be applied to program and control the behavior of multi-model LLM systems to maximize efficient use of the expanding and evolving language model ecosystem.
\end{abstract}

\section{Introduction}

The introduction of the transformer architecture and the self-attention mechanism has led to a an explosion of machine learning models for NLP and vision with a wide variety of model architectures, model sizes, varying sparsity, training paradigms and training data sets \cite{brown2020language, devlin2018bert, openai2023gpt4, he2022masked, taori2023stanford, bommasani2021opportunities, brown2020language}. Today the Huggingface (HF) model repository contains $\sim$262,000 machine learning models across language, vision and audio trained on tasks including masking,  distillation, instruction, Q\&A, PEFT, sparsification, quantization and RLHF. The HF model set includes large general purpose foundation models like LLAMA, Falcon variants; text2text models T5, FlanT5, Bloom, zero-shot classifiers, Bart-MBPI as well as smaller more focused models BERT, RoBERTa, ClinBERT, 
 PatentBERT, CodeBERT \cite{bommasani2021opportunities, hoffmann2022empirical, openai2023gpt4, dosovitskiy2020image, devlin2018bert} . Each variant commonly comes with sizing options such as tiny, base, large and X$^n$L. Users face an enormously context-dependent task in the selection and optimization of models for specific work-flows that might incorporate multiple tasks and domains. Thus, there is an urgent need for machine learning frameworks that can eliminate the burden of model selection and customization and unleash the incredible power of the vast emerging model library for end users.  

Currently, models are primarily compared with one another based upon leaderboards (eg Hugging Face, Dynabench, PapersWithCode) that contain summary statistics on benchmarking data and tasks (Squad Q\&A, Pile, GLUE). Engineers will select models based upon leaderboards, but then perform time-consuming testing and model integration to deploy a model into production. Summary statistics provide only a rough guide for users who aim to deploy models on data sets that might incorporate many different tasks and disparate data domains. Aggregate statistics are not representative of model responses on individual user prompts. A finance professional might need to parse financial text but also patents and legal data. A medical note contains both information on patient case history but also social history, medication and diagnostic information, and billing. In production, a model might need to perform a range of tasks (distillation, Q\&A, summarization)  on prompts drawn from distinct domains. Additionally, leaderboard based model selection can be applied once at model-selection-time but is expensive to maintain dynamically in production as the model ecosystem evolves while data sets and user needs might change continuously. The simplest solution of maintaining and deploying multiple LMs for a single task can be prohibitively expensive.

Further, users may want to perform a kind of Pareto optimization where they select models while considering additional goals and constraints including computational cost of a given model (model size), hallucination probabilities, recency, security, verbosity \cite{openai2023gpt4} etc. A user might prefer to decrease model size by $50\%$ (deploying a ‘tiny’ variant) if the decrease only incurs a $2\%$ change in accuracy on the ultimate user task, for example \cite{frankle2018lottery}. In fact, an organization might want to explore the entire cost/performance performance front of a model and dynamically move along this front depending on stringency and scale of a task and even real-time computing loads and GPU cost/availability. Edge applications of LMs on mobile phones, watches, and other IoT devices may not require high cost-high latency queries to LLMs. More generally, it’s not obvious that BERT-large, for example, will always perform better on every task than BERT-base or BERT-tiny; or that a code model will have optimal performance on a data set that contains code and also human read-able comments. 

Here, we introduce a perceptive routing architecture, Tryage (Figure \ref{fig:Fig1}),  inspired by the computational architecture of the human brain as well as Q-learning paradigms and the contextual bandit problem in reinforement learning. In Tryage a ‘perceptive router’ automatically routes user prompts to expert models within a model library through dynamic, on-the-fly analysis of the user problem as well as user supplied input constrains. In contrast to other emerging multi-model frameworks like Gorilla and Langchain that route based on analysis of a model card, the perceptive router in Tryage is itself a language model that learns through pre-training to predict down-stream, expert model accuracy on single prompts. The router, then, performs zero shot routing by integrating user flags through a single routing objective function.  The routing agent can integrate predicted accuracy with secondary user goals including model model size, model source, and makes dynamic routing decisions to down-stream expert models. 

We demonstrate that the Tryage system achieves expert level performance on individual prompts across domain specific data sets therby achieving an aggregate performance superior to any individual model and exceeding the performance of Roberta by as much as $17.9\%$ on specific domains and also outperforming Gorilla and GPT 3.5 turbo on model selection \cite{patil2023gorilla}.   Gorilla, in contrast to Tryage, makes routing decisions based upon analysis of model cards, descriptions of models, not based on quantitative predictions of model performance learned through training. By incorporating user constraints, Tryage enables users to incorporate constraints, for example, asking the router to prefer smaller (models having a lower number of parameters) allowing users to explore Pareto fronts of accuracy, model size, and other performance metrics (such as latency, reliability etc.).  In addition to performance, the router learns interpretable latent representations of data drawn from disparate domains. We anticipate routing models to be a broadly useful paradigm to enable distributed task execution and problem solving using sets of large language models. 

The Tyage architecture is inspired by the architecture of the human brain where the thalamus performs contextual routing of sensory inputs to 
cortical regions \cite{sherman2016thalamus,rikhye2018thalamic,wolff2019cognitive}. In the brain, the thalamic routing architecture supports modular processing of sensory data and problem solving  where tasks are broken down and routed to specialized expert processing units solve specific problems, including object identification, motion detection, object localization, and face recognition. Expert outputs can, then, be reassembled and presented to `higher-order' decision making systems \cite{mumford1991computational,mumford1992computational}. Mathematically, we are performing routing
using principles hypothesized to apply in neural routers in Tryage where the Tryage router builds an implicit model of down-stream experts models, predicting their behavior (accuracy on a given prompt), and then selects down-stream models based upon prompt specific performance estimates. We hypothesize that routing and modeling models can be widely incorporated into multi-model architectures to decompose tasks and integrate information from expert models into global decisions.





\section*{Tryage model architecture and mathematical framework}

\begin{figure}
    \centering
    \includegraphics[width = \columnwidth]{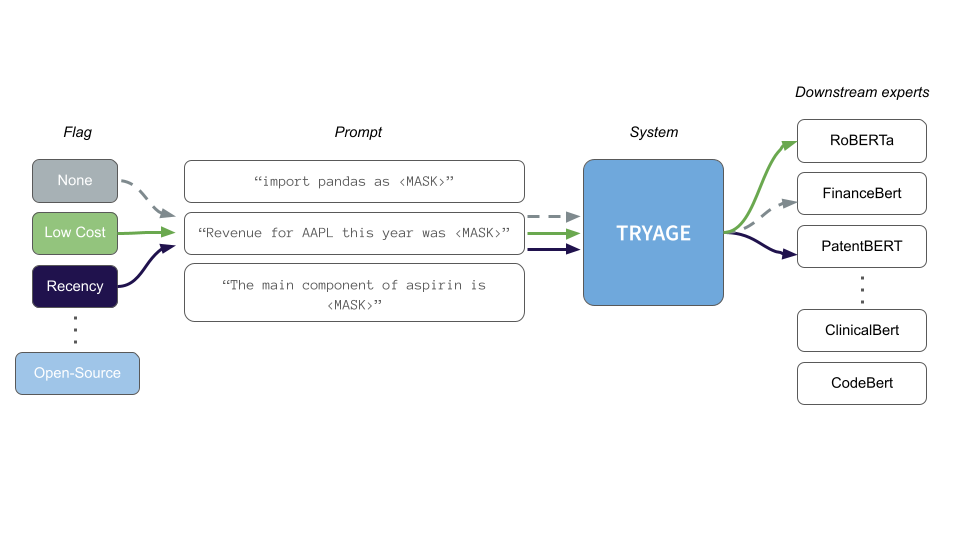}
    \caption{In the Tryage system, a prompt and flag are provided to the system, and it finds the best model to perform the MLM task given the flag and the prompt.}
    \label{fig:Fig1}
\end{figure}

\begin{figure}
    \centering
    \includegraphics[width = 0.8\textwidth, angle = 270]{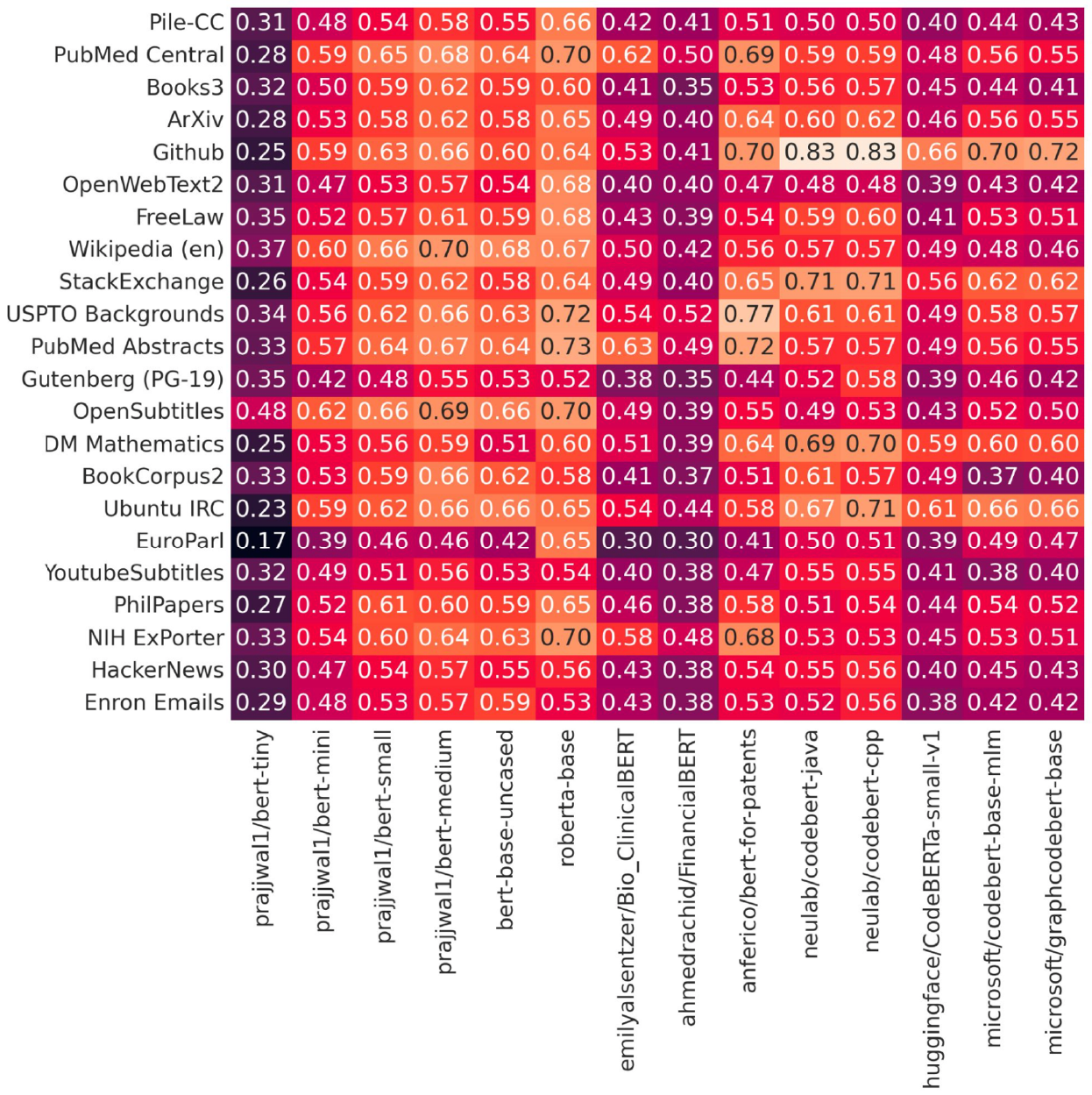}
    \caption{Picking a model for a task is challenging because multiple models display differential performance on different datasets. }
    \label{fig:Fig2}
\end{figure}

Language models are trained on data sampled from both generalized data sources like CommonCrawl as well as domain specific data (Pile sub-sets, financial data, clinical data), and also trained and fined tuned using different paradigms (Masked Language Modelling, Causal Language Modelling), and therefore, existing language models exhibit differential performance on down-stream tasks and data drawn from domains. Figure \ref{fig:Fig2} shows the performance of a set of popular transformers on masked language modeling across a series of distinct data domains. While the Roberta model exhibits the highest mean accuracy across all data domains in the Pile data sets, expert models like Codebert and Patent Bert significantly outperform Roberta when input data is drawn from specific domains like GitHub or a patent database. The empirical results motivates the development of a multimodel architecture that can dynamically select optimal models depending on a specific input. The goal of the system is to maintain high general accuracy while capturing expert level performance when prompts or tasks are drawn from specific data distributions. 

To enable dynamic routing of user prompts and tasks to a library of models, we design a perceptive routing system. The system contains a Tryage (routing) node which executes a routing function and a library of down-stream/leaf nodes (Figure \ref{fig:Fig1}). The Tryage node accepts a user prompt and selects a down-stream model based on the prompt. The routing function combines the performances of the downstream models and integrates their predictive accuracies with user supplied constraints/goal flags into a single objective function. The routing function selects a specific model by trading off predicted accuracy against a set of parameter weighted constraints. Optimization of the routing loss function leads to a model selection decision, and the prompt is routed to the downstream model. 

In the following section we describe the individual components of the Tryage system in mathematical detail including Tryage model training and the mathematical structure of the routing objective function

\subsection*{Preliminaries and Notation}

Consider a language model library $\textbf{M} = (M_1 \dots M_n)$  where $M_i$ are individual models. Each model $M_i \in \mathbf{M}$ takes input $z$ and maps the input to an output $M_i(z)$ which, for example could be next word prediction probabilities. In general, different elements of $M$ will have differential performance on a prompt and tasks due to model architectures, training data, and training paradigm. Our goal is to develop a model selection/prompt routing strategy in which a routing model $R$ can select the optimal expert model $M_i$ for a specific prompt while incorporating user constraints as secondary loss functions into the routing decision. A naive strategy might select down-stream models based on data set labels (e.g. select a clinical model for analyzing a clinical trial report). However, in applications, users often seek to analyze data streams that contain information from multiple domains, and even a file of python code might contain code and comments; a clinical trial report will contain biomedical and regulatory data. Further, in applications, data labeling is laborious and expensive, and so user data is often unlabeled. Therefore, our goal is to select an optimal expert model through dynamic analysis of individual prompts $z$. Given such a framing, the model selection problem has conceptual and mathematical similarities with problems in reinforcement learning (RL) including the multi-armed bandit and conceptual bandit problems, and we draw inspiration from RL strategies for optimal action selection including Q-learning. 

In Tryage, given an input $z$, and a task with loss function $L(z, M_i)$ taking as inputs the user prompt $z$ and a model $M_i$, we construct a $Q$ function mapping actions, $\pi_i$ to rewards, $r_i$. Specifically, the actions represent the routing of a prompt $z$ to each possible expert models $M_i \in M$.  In our general implementation, $Q(z,\pi_i)$ will estimate the reward, $Q(z,M_i) = \hat{L}(z,M_i,)$, achieved by model $M_i$ on prompt $z$, and the Q-function will be implemented through a language model, itself trained to predict down-stream loss values. The Q-function sends estimated loss values to a routing function that, then, makes a routing decision by combining loss estimates with goals/constraints implemented through weighted constraint functions. 

\subsection*{Oracle Router Model}

To describe the routing decision, we initially consider a theoretical (but impractical) Oracle Router Model where a router model has access to the entire $Q$ table of performance metrics (loss values) for each of the expert models on all prompts. The router model can then select the model that maximizes performance while incorporating secondary user goals in the form of constraint functions $C_j(M_i)$ with weights $\lambda_j$. Given complete knowledge of Q, the router selects the model through optimization of a function that makes trade-offs between loss and constraints $C_i$

\begin{equation} \label{eq:router}
\begin{split}
    \hat{M}_O = R_O(z, M; \lambda) & =  \underset{M_i}{\arg \min} \Bigg[ \ Q(z,M_i) + \sum_{j} \lambda_j \ C_j(M_i) \Bigg] \\
  &=  \underset{M_i}{\arg \min} \Bigg[ \ L(z, M_i) + \sum_{j} \lambda_j \ C_j(M_i) \Bigg]
\end{split}
\end{equation}

Where $L(z,M_i)$ is the loss of model $M_i$ on prompt $z$, $C_j(M_i)$ represents a scalar valued constraint function $C_j$ applied to model $M_i$ (for e.g., a model size penalty such as $\log(\# \text{parameters})$ and $\lambda_j$ are weight parameters that allow the user to weight different user constraint functions. The routing objective function allows the routing model to trade-off accuracy goals against constraints. For any setting of $\lambda_j$, the router model is able to combine the Q values and constrains values, and pick the model that minimizes the combined loss. We note that in Tryage, user constraints are incorporated into the input prompt itself, for e.g. “The capital of California is [blank] [Flag: Smallest model]”. These two costs combine to give the routing loss $L_R$, which the Oracle Router  model $R_O$, minimizes to output a predicted model $\hat{M}_O$. 

Given a Q oracle, the routing objective function can be minimized over $Q$ and $C_j$.
Minimization can be performed for distinct values of $\lambda$ for example modulating the weighting of model size from unimportant $\lambda \approx 0$  to dominating $\lambda > 0 $. The routing function can incorporate multiple objective functions by incorporating constraint functions $C_j(M_i)$ that score each $M_i$ for factors including recency, verbosity, hallucination probability etc.

\subsection*{Predictive Router model}

While an oracle strategy might select down-stream models based on supplied data, it is computationally inefficient to create a Q table for each incoming prompt by running all $|M|$ models on all possible prompt instances. Therefore, our goal is to train a Q function that estimates down-stream loss, so that the function can return an estimated loss through dynamic analysis of individual 
input prompts. One way to analyze the prompt locally is to predict the Q table by training the routing model in a supervised learning paradigm on a data sets of prompts and expert model losses. 

We use a supervised training paradigm where we construct a differentiable function $R(z,M_i; W)$ with weights (model parameters) $W$ where $R$ takes a prompt $z$ as input and returns a scalar in $\mathbb{R}$ of estimates loss values for model $M_i$.  For $R$ we can use any function differentiable in $W$. In practice we implement $R$ using an existing  language model as the base-architecture, and modify the architecture so that $R$ outputs an $n$ dimensional output vector over the real numbers. For example, we achieved favorable loss prediction accuracy with Bert-tiny. In general, given an architecture, we train $R$ via 

\begin{equation} \label{eq: predictive router}
\min_W \ \ \mathbb{E}_{z \sim p(z)} \ \left( \frac{1}{|M|} \sum_{M_i} D\big(R(z,M_i;W)||L(z,M_i)\big) \right)
\end{equation}

where $D(\cdot||\cdot)$ is divergence between predicted loss $R(z,M_i;W)$ and ground truth loss $L(z,M_i)$ for prompts $z$ drawn from data distribution $p(z)$. Expectation is taken as empirical expectation and divergence is summed over all models $M_i$ in the model library. 

We train the predictive router by solving equation
\ref{eq: predictive router} using stochastic gradient descent over batches of size $m$ as

\begin{equation} \label{eq: gradient descent}
 \nabla_W  \ \frac{1}{m} \  \sum^m_{j=1} \sum_{i=1} ^{|M|} D\big(R(z_j,M_i;W)||L(z_j,M_i)\big). 
\end{equation}

In general, we were able to train router models that approximate loss within $\epsilon = .1$ of true loss for masked language modeling. 

While we do not explore the implementation here, we also note that the entire architecture including the router and down-stream models can be trained end-to-end. Specifically, the router selects the model by minimizing the function shown in the RHS of equation \ref{eq:predictive_router}.  $\hat{M}_P$ is the model predicted by the predictive router, $\hat{L}(z, M_i)$ is the router model prediction of $M_i$'s performance on prompt $z$.

\begin{equation} \label{eq:predictive_router}
\begin{split}
    \hat{M}_P = R_P(z, M, \lambda) & = \underset{M_i}{\arg \min} \ \hat{L}(z, M, \lambda) \\
    & = \underset{M_i}{\arg \min} \Bigg[ \ \hat{L}(z, M_i) + \sum_{j} \lambda_j \ C_j(M_i) \Bigg]
\end{split}
\end{equation}

Now, we also update the expert model $\hat{M}_P$ 

\begin{equation}
\displaystyle  \min_{W_{\hat{M}_P}} L(z,\hat{M}_P, W_{\hat{M}_P})
\end{equation}

by traversing $\nabla_{W_{\hat{M}_P}} L(z,\hat{M}_P;W_{\hat{M}_P})$ where we are now updating the weights of model $\hat{M}_P$ to decrease loss on the prompt $z$ sent by the routing model. In this way, the routing model and expert models can be trained together so that each expert model can increase performance on the specific set of prompts being sent to it by the router. The paradigm has conceptual similarity to the training of self-organized maps. In each training step, we update the routing model $R$ as well as the expert models $M$, so that each expert model develops its own expertise. We train the whole system end-to-end, and decouple the updates to the router and expert models to make the training more modular.

\section*{Experiments with Tryage}

The Tryage architecture is designed to utilize a family of models to achieve expert level performance on NLP spanning different data domains in a single architecture. To test and demonstrate the performance of Tryage, we trained a Tryage model for model selection on a library of $11$ natural language models drawn from the Hugging Face ecosystem. We benchmark the performance of the resulting model architecture for accuracy against model selection performed by Gorrilla and GPT3.5 across data domains drawn from the Pile. We also show that through training, the Tryage model system (router plus experts) develops emergent capabilities including the ability to generate unsupervised latent representation of data across domains that are more natural and interpretable than GPT2. Finally, we demonstrate adaptive constraint based routing and compute a Pareto front of optimally for global size/accuracy tradeoffs.

\subsection*{Tryage outperforms Gorrila on unconstrainted model selection for MLM on Pile despite being 1000x smaller}
To benchmark, we trained a Tryage system to perform masked language modeling on the Pile data set \cite{gao_pile_2020} and compared performance with state of the art perceptive model seletion systems. We use MLM as a benchmarking task because of its objective evaluatability, wide model availability, and its base as a paradigm for training large foundation models. The  data set is a standard multi-domain, extremely large repository of textual data. The  incorporates both generic text like common crawl but also niche domains like Gutenberg, GitHub, Stack Overflow, USPTO patents, etc. As the routing model, we selected BERT-small since initial experiments suggested that larger models did not yield better performance. As the downstream expert models, we chose $11$ variants from huggingface including ClinicalBert, SECBert, FinancialBert, PatentBert, CodeBert and general Bert family models including Roberta, small bert variants, bert-base etc.

We trained the router as described on 10 million tokens per epoch using ADAM with a weight decay of $1e-5$ and a learning rate of $5e-5$ that we exponentially decayed by 0.9. To maximize the number of tokens exposed, we curtailed each input example to 512 tokens, and used a batch size of 24 per GPU, and trained our models on A100s with 80 GBs of RAM. We tried to prevent the Tryage model overfitting by using an early stopping criterion with a patience of 16, conditioned on the validation loss, where the validation loss was measured 4 times per epoch. We used model checkpointing to use the best performing model (best performing in terms of validation loss) to run over our test set. We trained using the pytorch lightning framework. 

Following training, we evaluate performance on test data from the Pile, and we found that Tryage achieved near expert model level performance across the data set, outperforming state of the art approahces like Gorrila.

\begin{figure}
     \centering
     \begin{subfigure}[b]{0.49\textwidth}
         \centering
         \includegraphics[width=\textwidth]{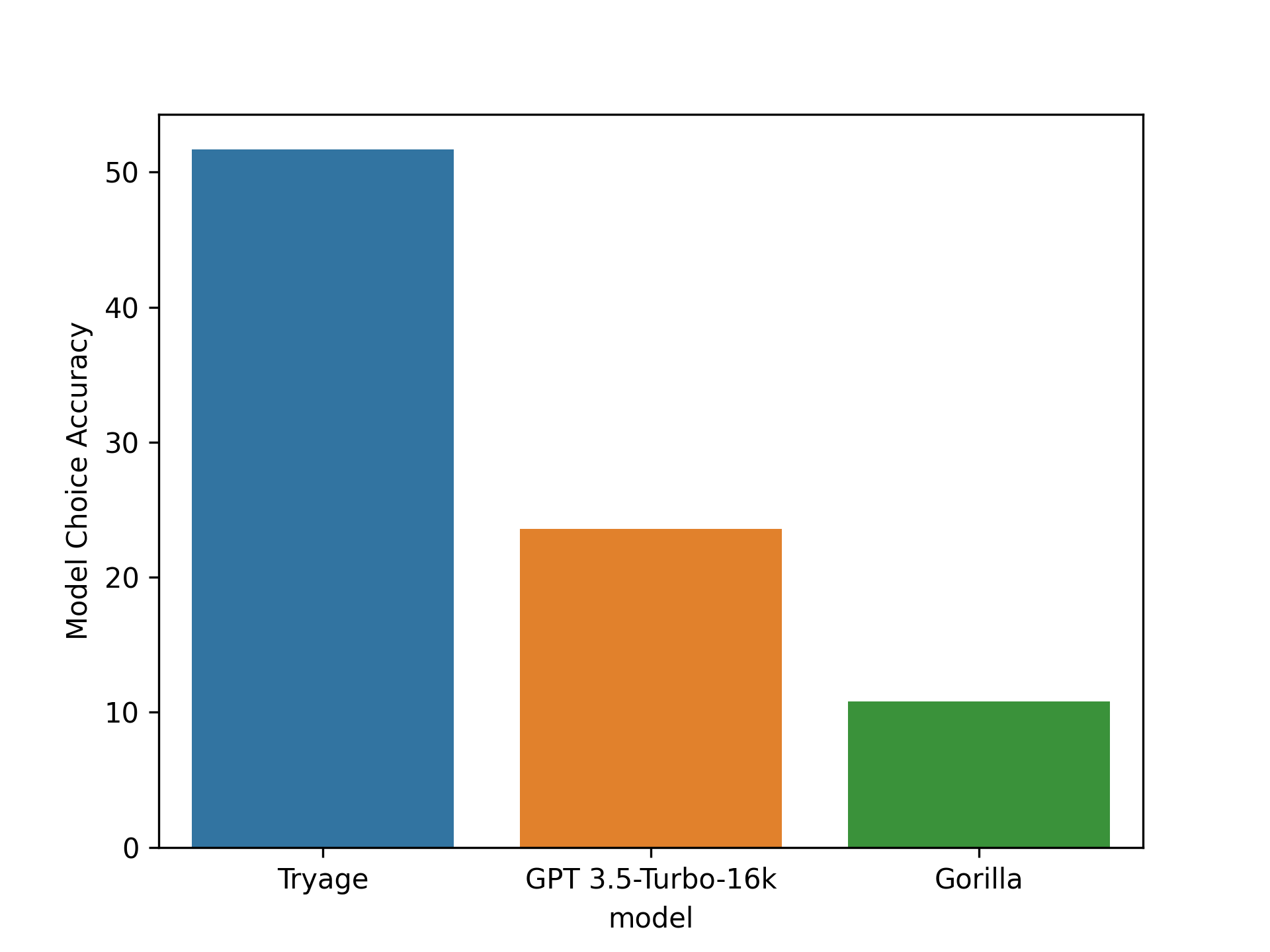}
            \caption{}
        \label{fig:model-cx-acc}
     \end{subfigure}
     \hfill
     \begin{subfigure}[b]{0.49\textwidth}
         \centering
         \includegraphics[width=\textwidth, angle = 270]{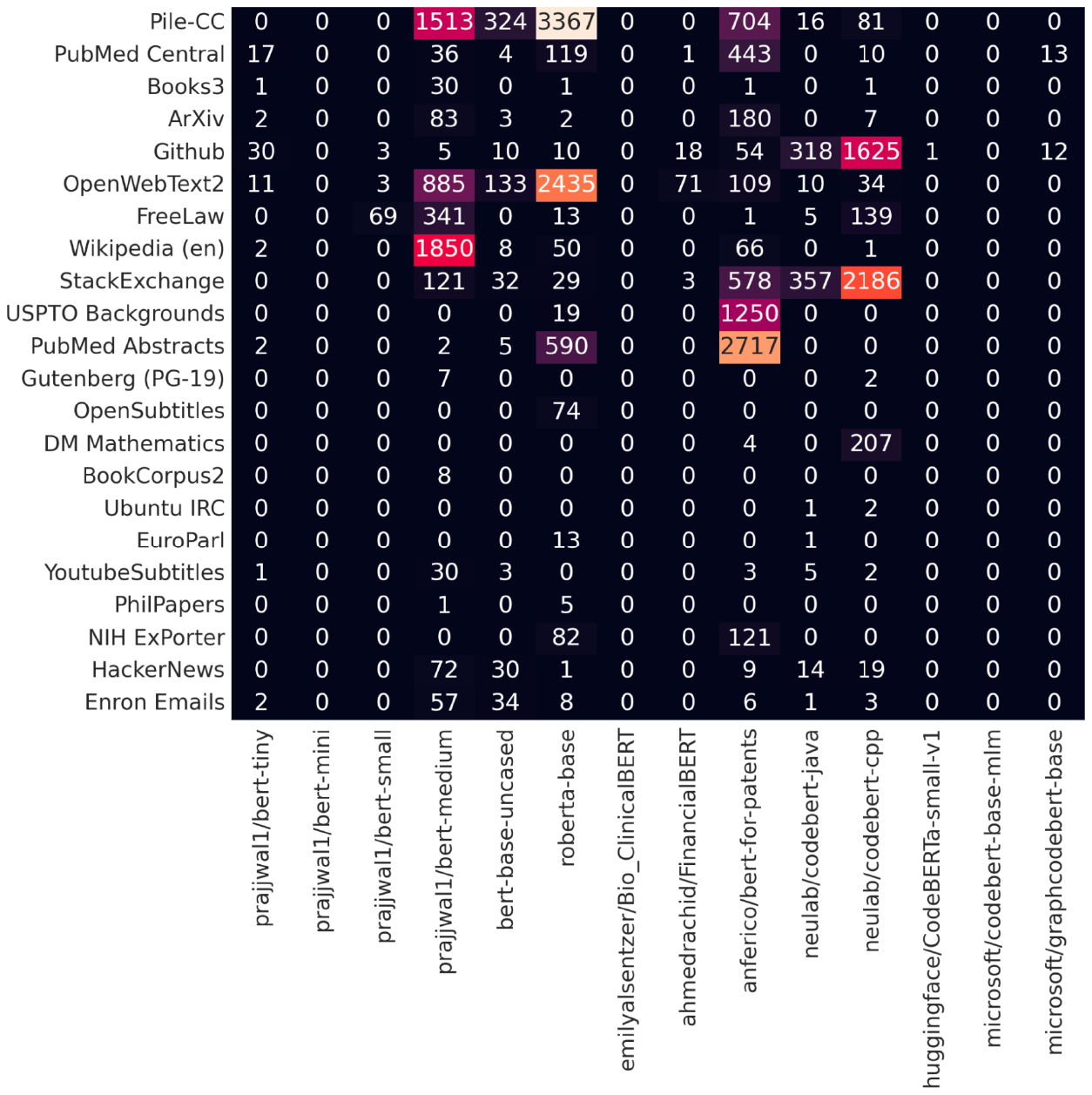}
        \caption{}
        \label{fig:model-cx-ll}
     \end{subfigure}
     \hfill
     \begin{subfigure}[b]{\textwidth}
         \centering
         \includegraphics[width=\textwidth]{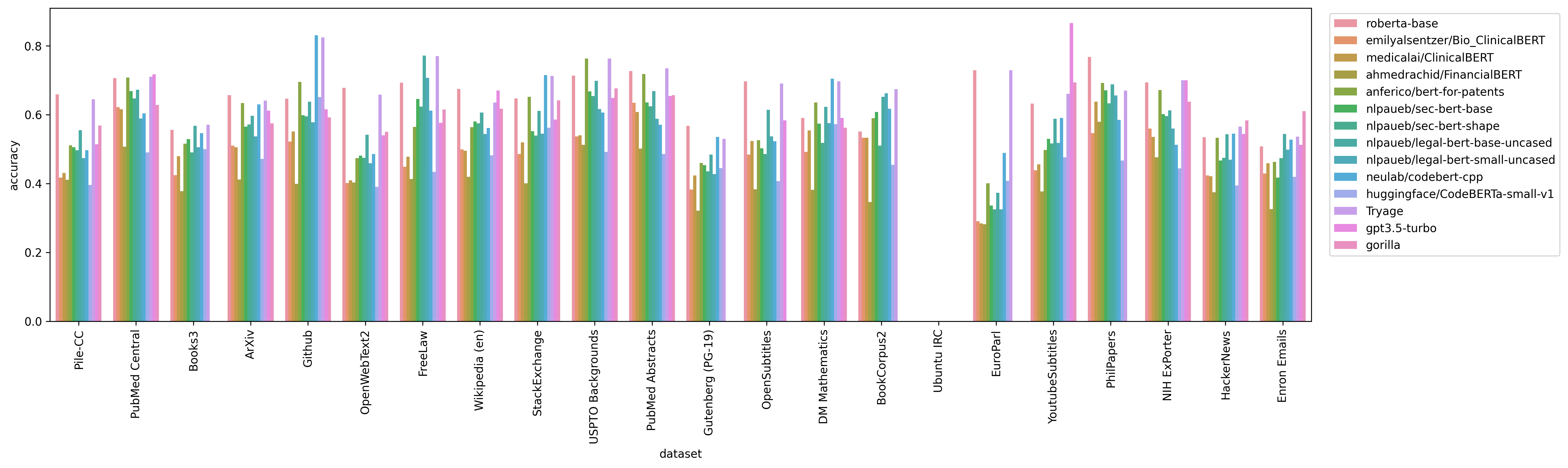}
         \caption{}
         \label{fig:full_performance}
     \end{subfigure}
     \hfill
     \begin{subfigure}[b]{\textwidth}
         \centering
         \includegraphics[width=\textwidth]{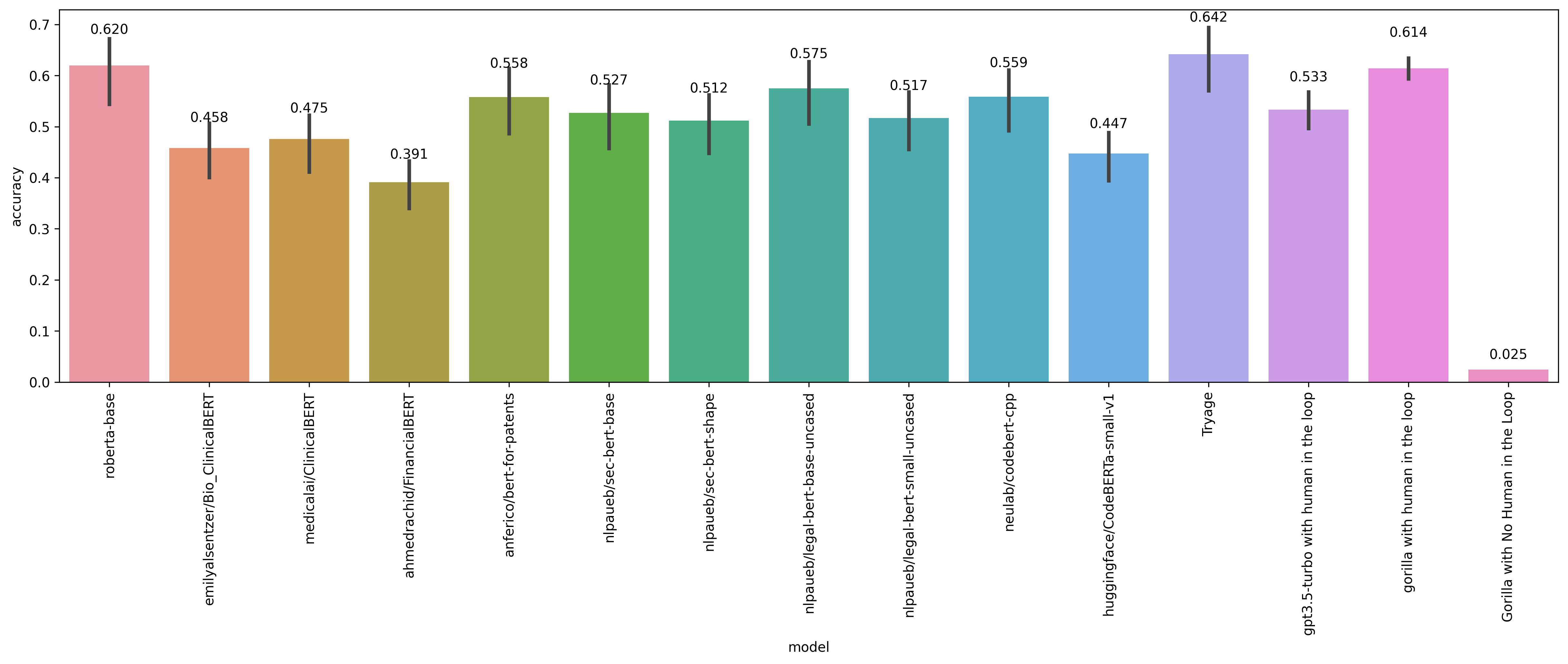}
         \caption{}
         \label{fig:accs_agg_all_models}
     \end{subfigure}
        \caption{a) - Tryage outperforms existing SoTA models on a task of Masked Language Modeling including SoTA generative LLMs such as GPT 3.5-Turbo and Gorilla. b) - By examining which models the tryage system picked for incoming queries of each domain, it becomes easier to build a super-specialized system that can still generalize with high performance across multiple specializations. It also helps build trust and transparency in the system, since one can use tryage to filter for specific domain knowledge of a certain dataset. c) - Performance of models in the MLM task across all the subsets of the  shows that the Tryage system outperforms other models trained on MLM. d - Same as c, but averaging across datasets by model, shows gains by Tryage, as well as performance of Gorilla without a human in the loop}
        \label{fig:Fig3}
\end{figure}

We found that Tryage was able to recognize the ideal model to send the prompt to with an accuracy of 50.8\% of the time, compared to 23.6\% by GPT 3.5 Turbo and 10.8\% by Gorilla (\ref{fig:model-cx-acc}) \cite{patil2023gorilla}. The numbers for GPT and Gorilla are overestimates, since we searched for whether the generated output contained `any' evidence for the best model, rather than `only' evidence for the best model. 

With Gorilla, we observed that the code generated by Gorilla compiled only 23\% of the time. We found however, that the model did not produce outputs requested any of the times. Gorilla did produce outputs that could be used by a human about 10\% of the time, but required a lot of post-processing to coerce into the requested form. We attempted a Gorilla-with-human-in-the-loop interpretation of the Gorilla results. We scanned the Gorilla output for any evidence of it using one of the models from our list, and averaged the MLM accuracies of the selected model for each example in our dataset. We found that the average MLM accuracy was 61\%, lower than Roberta-base or Tryage, despite being over 1000x (larger Figure \ref{fig:full_performance}, Figure \ref{fig:accs_agg_all_models}). Further, only 50\% of the times did Gorilla actually use a model that was in the list of models we specified in the prompt. 

\subsection*{Tryage achieves interpretable routing of data to models}

Following training, we analyzed routing of specific prompts to down-stream expert models based upon their data domain. We found that the tryage model correctly sends prompts from code datasets such as Github and StackExchange to code models, USPTO Backgrounds and Pubmed Abstracts to patent models and CommonCrawl and OpemWebText2 to general english models (Figure \ref{fig:model-cx-ll}). Specifically, 78\% of the examples taken from Github were sent to a C++ model, 98\% of examples from DM Mathematics were sent to the C++ model and 98.5\% of data from the US Patent and Trademark Office (USPTO) backgrounds were sent to the patent model.

\subsection*{Tryage matches expert level performance on mission critical, domain specific datasets}

Large gains of up to 17.9\% were observed on code, math and patent datasets, where Tryage was able to correctly allocate to specialized models (full results shown in figure \ref{fig:full_performance}). Datasets (gains) that the Tryage framework was able to improve performance over and above RoBERTa, included Github (17.9\%), Freelaw (7.7\%), Stackexchange (6.5\%), DM Mathematics (10\%), BookCorpus2 (12.3\%) and USPTO Backgrounds (5.02\%).

\begin{figure}
     \centering
     \begin{subfigure}[b]{0.48\textwidth}
         \centering
         \includegraphics[angle = 270, width=\textwidth]{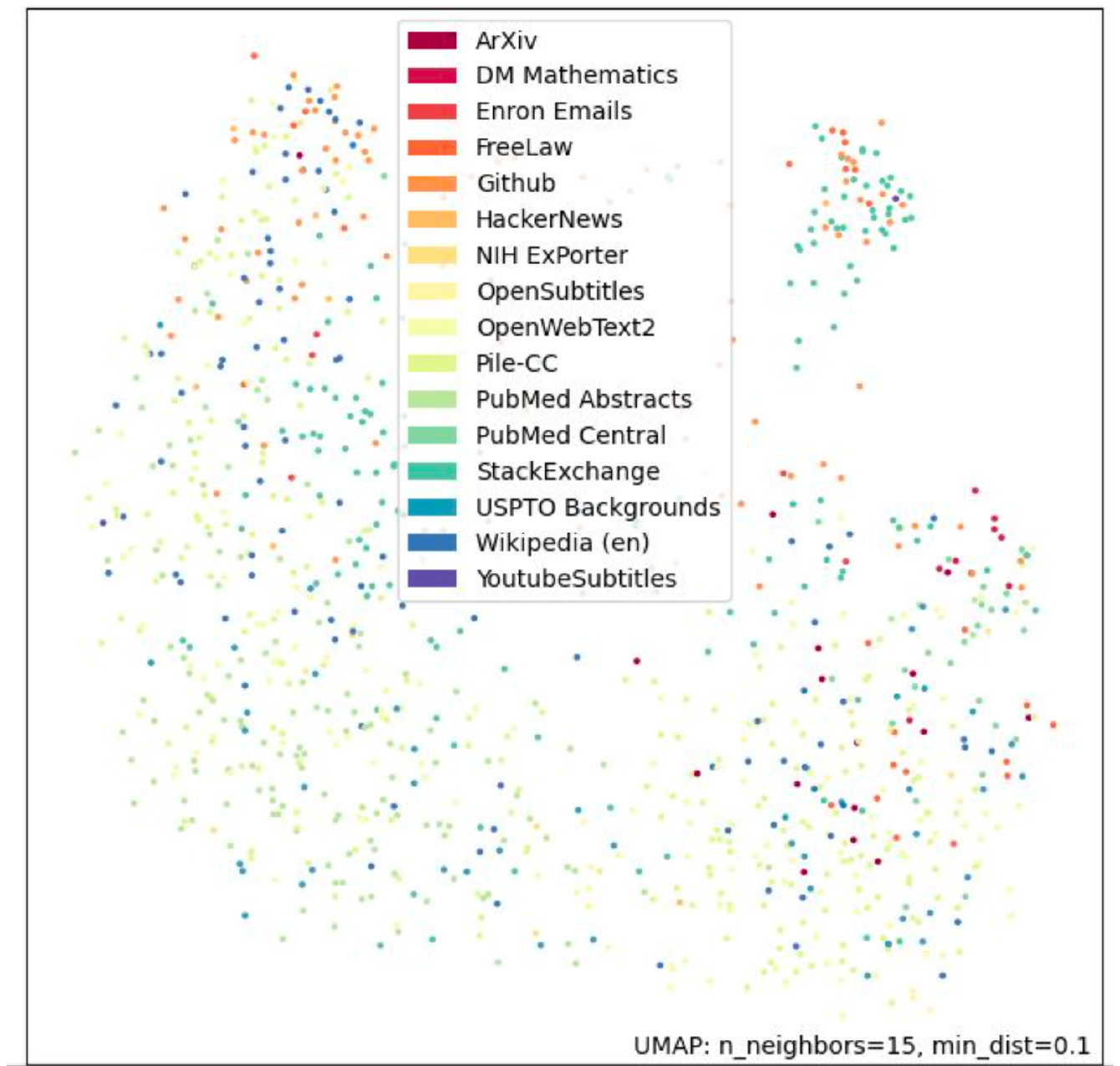}
         \caption{GPT UMAP}
         \label{fig:GPT_UMAP}
     \end{subfigure}
     \hfill
     \begin{subfigure}[b]{0.48\textwidth}
         \centering
         \includegraphics[width=\textwidth]{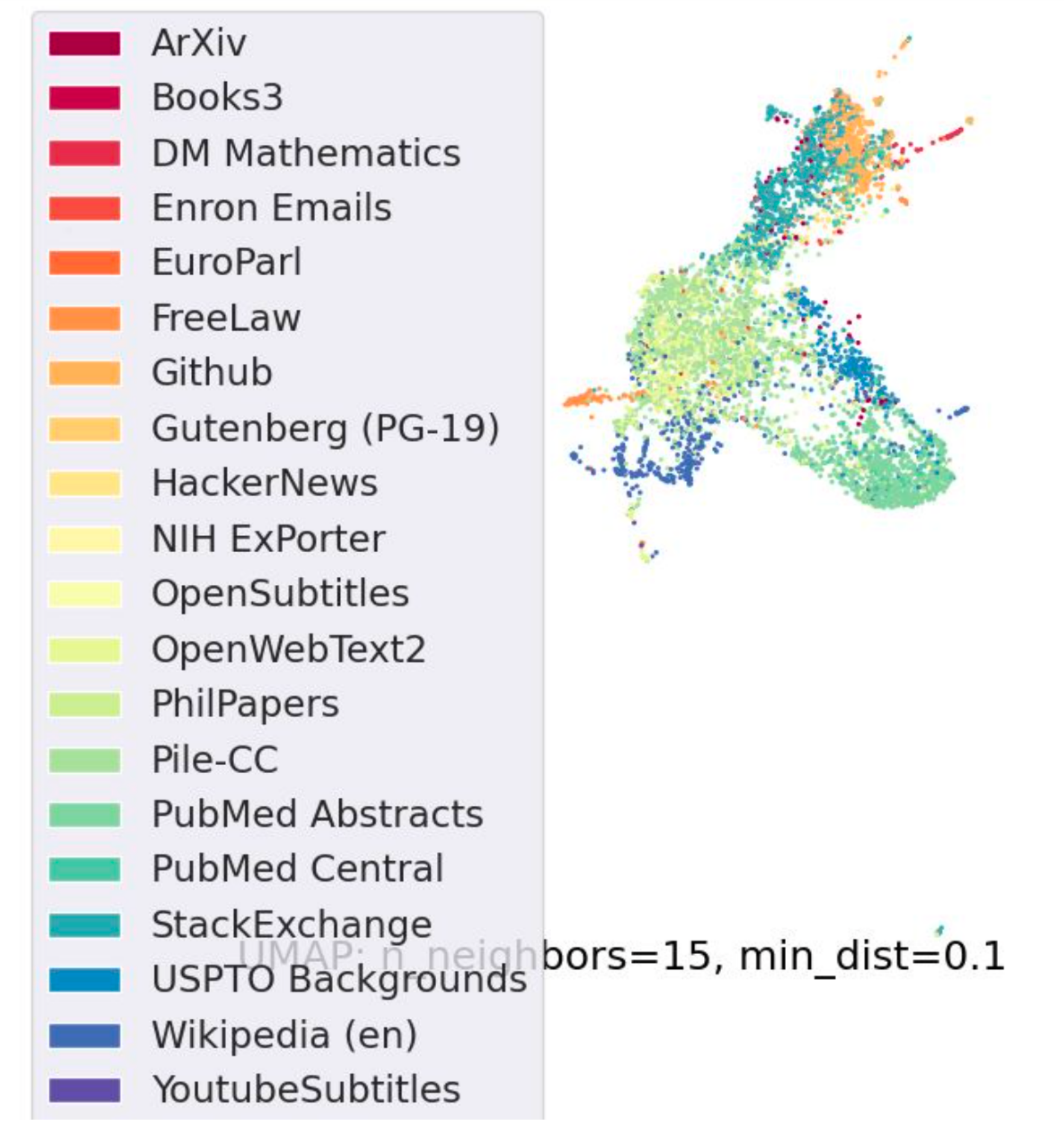}
         \caption{Tryage UMAP}
         \label{fig:Tryage_UMAP}
     \end{subfigure}
     \hfill
        \caption{By examining the UMAP of the latent space embeddings of queries coming into the tryage model (b), we observe the embeddings of a specific domain cluster. On the other hand, a general model like GPT-2, doesn’t display the same behavior (a) where the points of the same domain don’t cluster. }
        \label{fig:UMAP}
\end{figure}

\subsection*{Tryage generates latent separation of data into domains}

In addition to achieving expert model performance across domains, Tryage also constructs an implicit latent representation of language during training  without supervision (Figure \ref{fig:UMAP}). Specifically, one can see clustering of data coming from Github, CommonCrawl, Arxiv, StackExchange, PubMed and Wikipedia in the UMAP generated by Tryage (Fig \ref{fig:Tryage_UMAP}). In comparison, Tryage offers better latent space separation than GPT-2, whose embeddings are shown in a UMAP in Figure \ref{fig:GPT_UMAP}, since the embeddings generated by GPT-2, are diffuse and not clustered. The UMAP space with default parameters, provides interpretability since one can infer that the model has learned separations between the subsets of the data.

\begin{figure}
     \centering
     \begin{subfigure}[b]{0.8\textwidth}
         \centering
         \includegraphics[width=\textwidth]{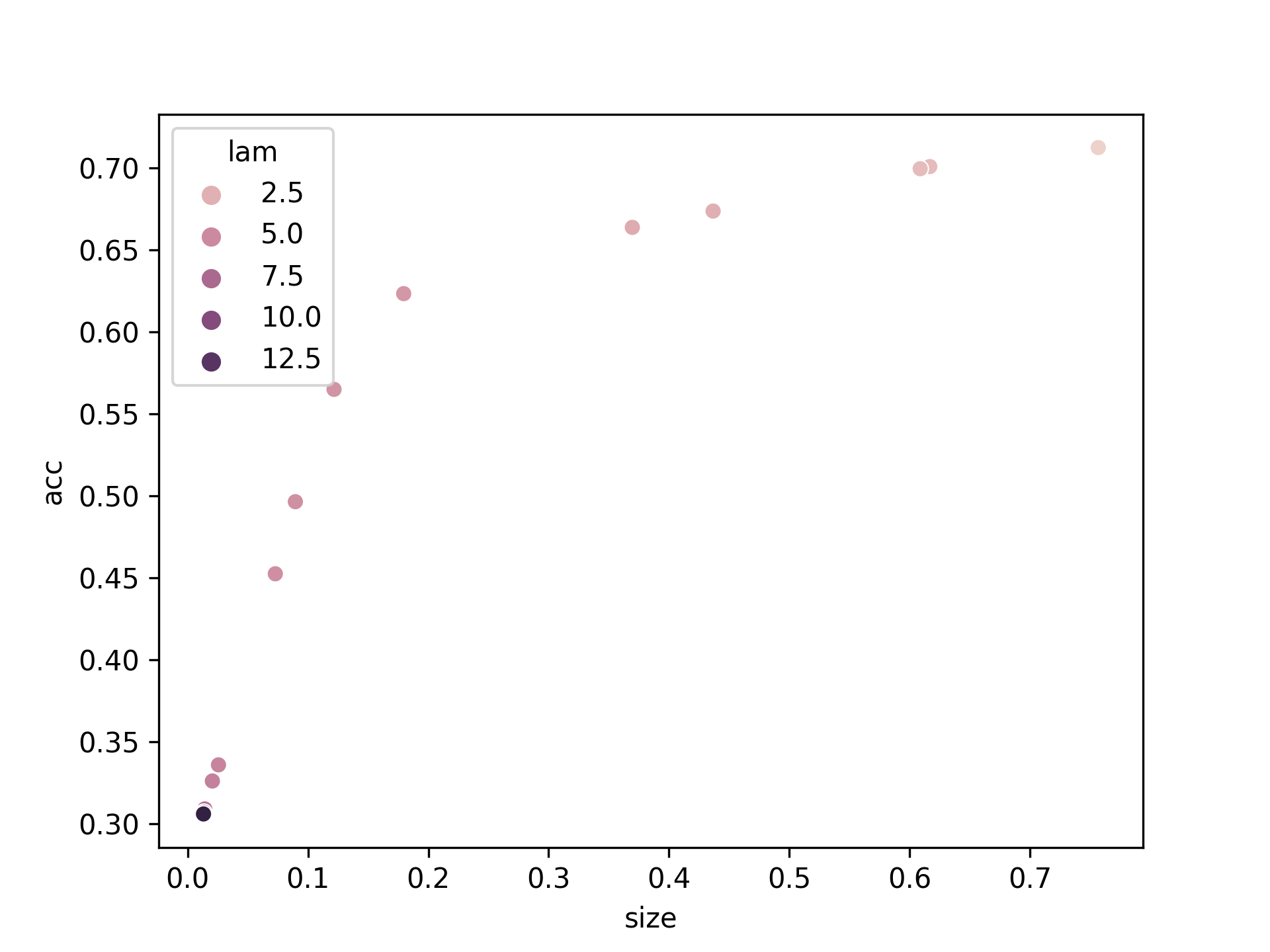}
         \caption{}
         \label{fig:lam-vs-perf}
     \end{subfigure}
     \hfill
     \begin{subfigure}[b]{0.3\textwidth}
         \centering
         \includegraphics[width=\textwidth]{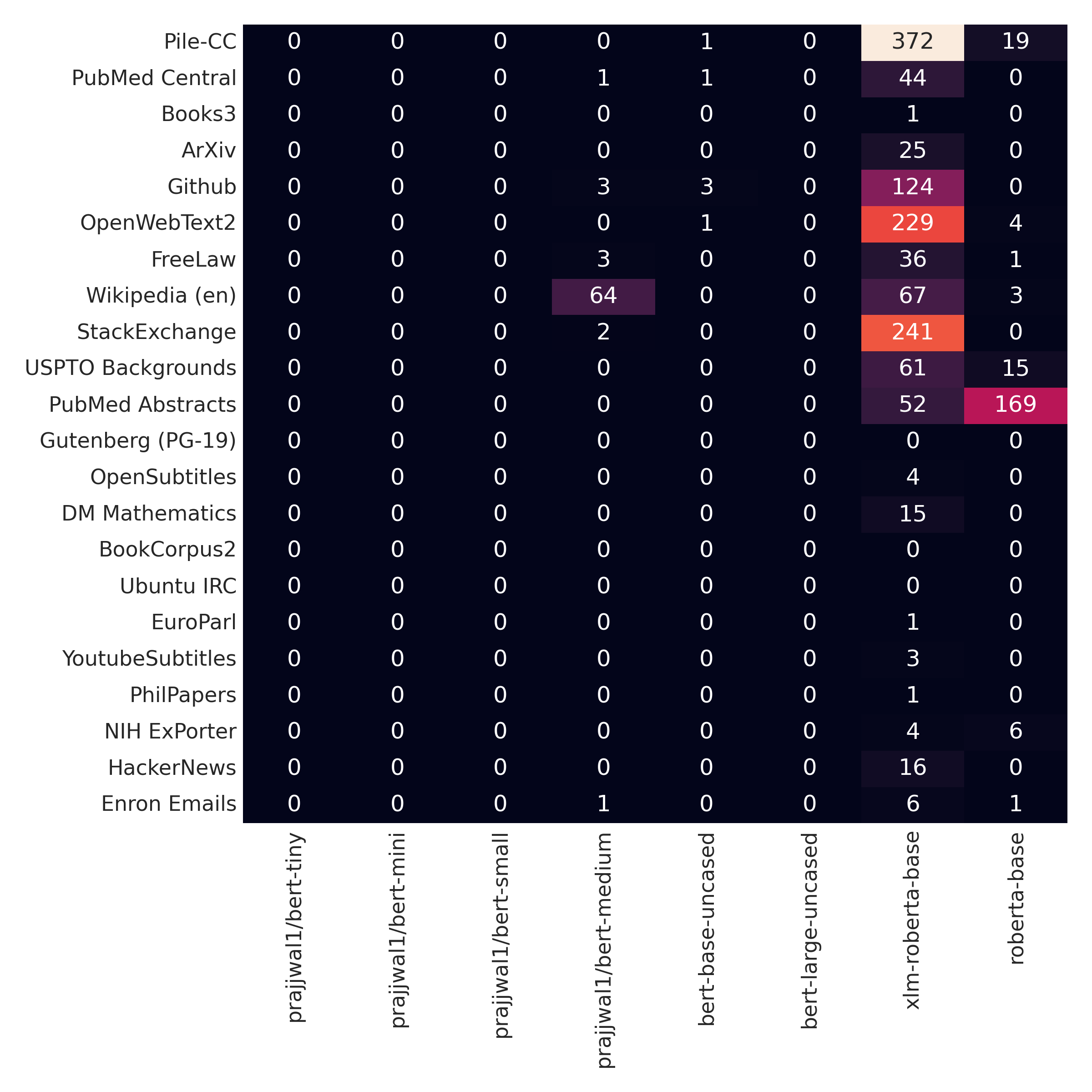}
         \caption{$\lambda = 0.33$}
         \label{fig:lam1}
     \end{subfigure}
     \hfill
     \begin{subfigure}[b]{0.3\textwidth}
         \centering
         \includegraphics[width=\textwidth]{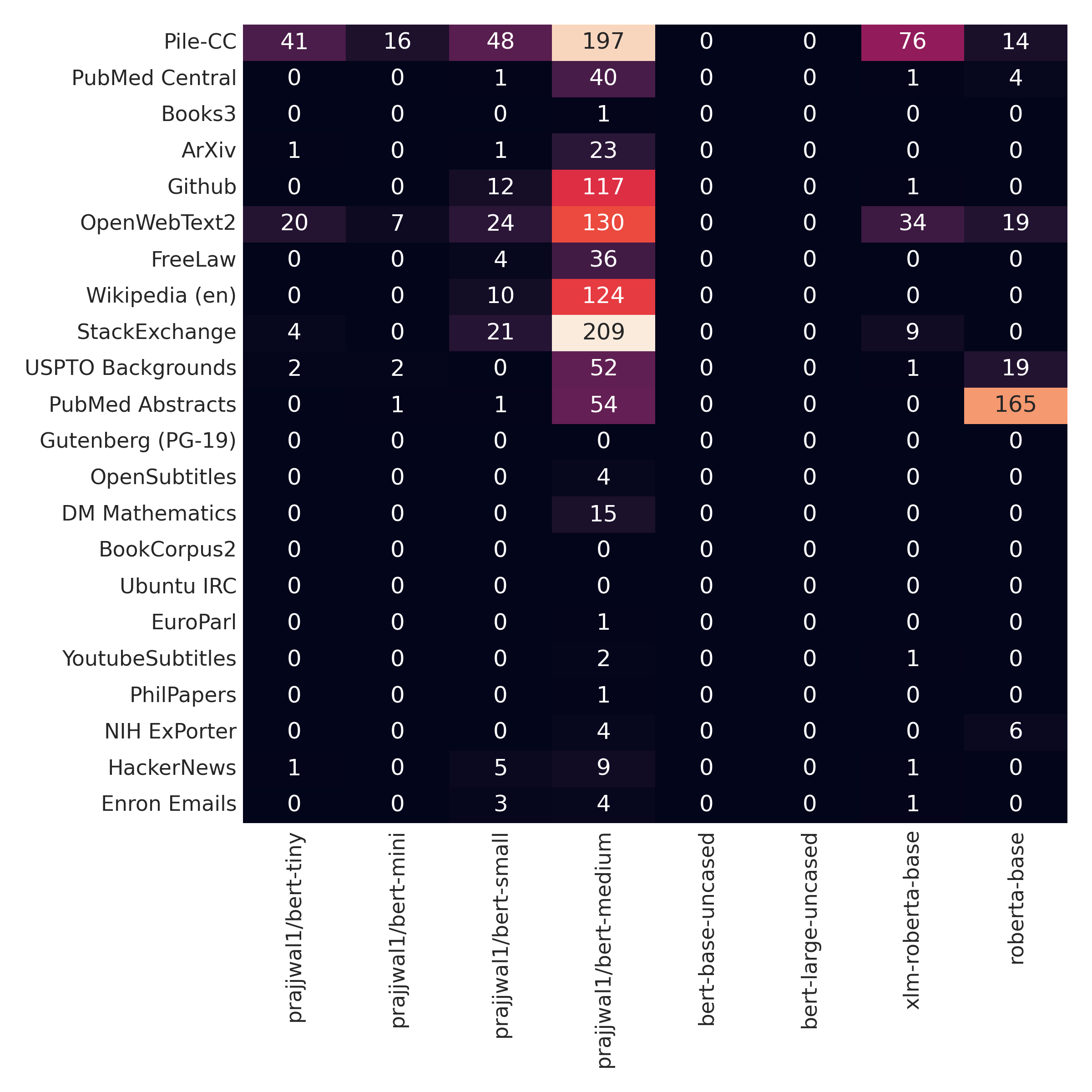}
         \caption{$\lambda = 4.2$}
         \label{fig:lam2}
     \end{subfigure}
     \hfill
     \begin{subfigure}[b]{0.3\textwidth}
         \centering
         \includegraphics[width=\textwidth]{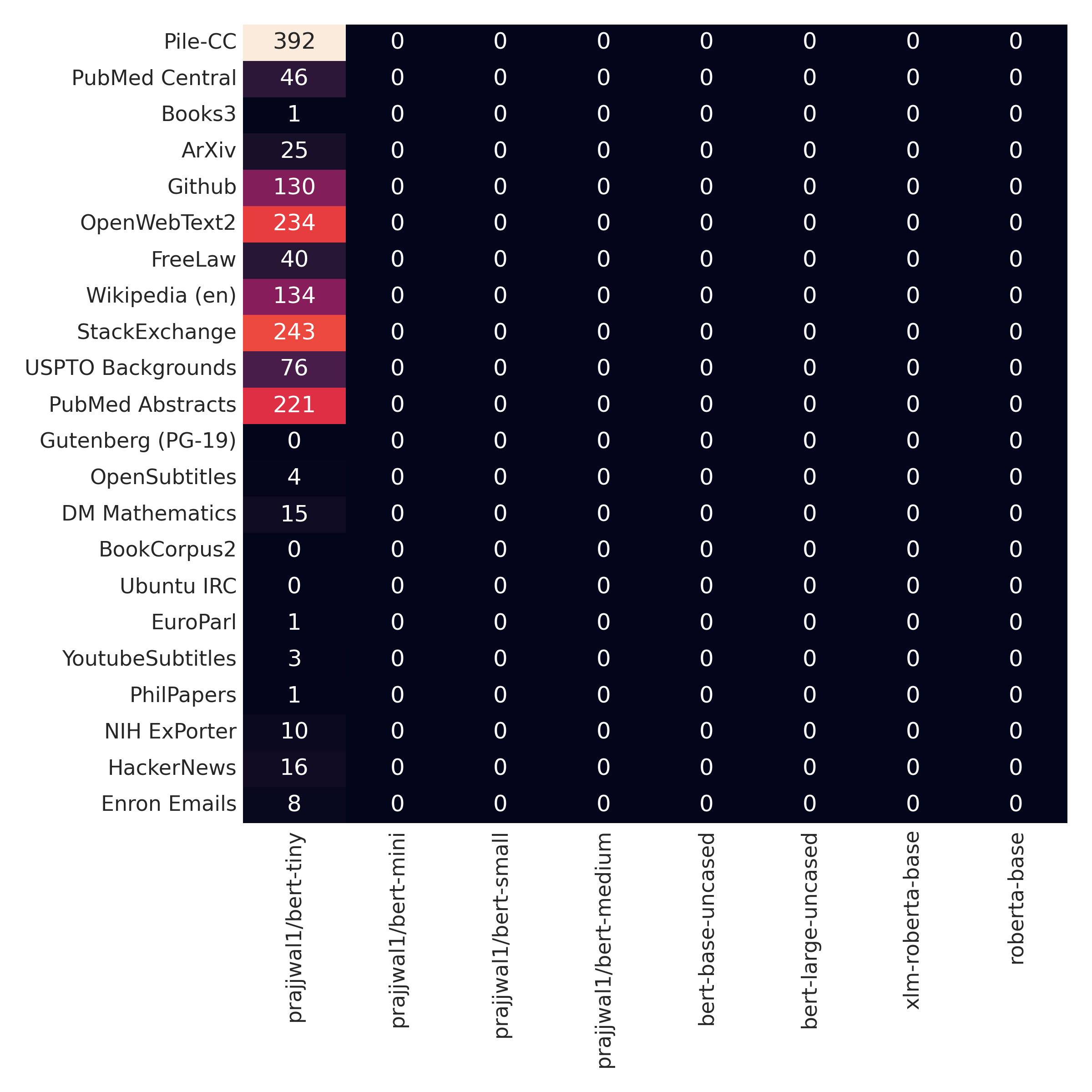}
         \caption{$\lambda = 13.9$}
         \label{fig:lam3}
     \end{subfigure}
        \caption{a) - By trading off performance for effective model size, we can form a Pareto curve along which users can choose their desired performance / latency preference. (b, c, d) - By examining the allocations of the tryage system while increasing $\lambda$, we can see the allocation change from predominantly to large models (b)  to a mixture of medium, large and small models (c) to predominantly smaller models (d). This allows for a system that only queries large models when it `needs’ to, effectively parlaying cost for performance.}
        \label{fig:Fig5}
\end{figure}



\subsection*{Tryage enables routing and model selection along Pareto front}

Users often seek to incorporate specific preferences and needs, for example, users might aim to prefer smaller less resource intensive models for tasks. Tryage enables weighting of models based on constraints through objective function (eq \ref{eq:router}). To demonstrate constrained routing, we added a linear model size penalty $C(M_i) = \frac{|W_i|}{\max{|W_i|}}$ (number params in model $i$ divided by max params in model library) that is weighted by a parameter $\lambda$. In our implementation, we set $\lambda$ as a hyper-parameter and search over $\lambda \in [0,2^4]$ and computed the net Tryage combined accuracy on MLM. The computation generates a Pareto front or trade-off curve enabling users to select values of $\lambda$ based on size accuracy tradeoffs. We note that in future releases we can tie $\lambda$ to a natural language prompt. 

We found that varying the size penalty, forced the choosing of smaller models. With only a 5\% reduction in combined accuracy (\ref{fig:lam-vs-perf}, the user was able to save over 50\% of compute. The gradual shift in allocation is evident in Figures \ref{fig:lam1}, \ref{fig:lam2}, \ref{fig:lam3} where an unconstrained Tryage system allocated mostly to larger models, and gradually tryaged to smaller models, as $\lambda$ was increased.

\bibliographystyle{plain} 
\bibliography{References/references,References/scibib, References/surya-zotero}

\end{document}